\documentclass[runningheads]{llncs}
\usepackage{graphicx}
\usepackage{comment}
\usepackage{amsmath,amssymb} 
\usepackage{color}
\usepackage{caption}
\usepackage{multirow}
\usepackage{cite}
\usepackage{hyperref}
\usepackage{algorithm}
\usepackage{algpseudocode}
\usepackage{authblk}

\usepackage{booktabs}
\usepackage[font=small]{caption}

\newcommand{\bx}{\mathbf{x}}
\newcommand{\tby}{\tilde{\mathbf{y}}}
\newcommand{\bw}{\mathbf{w}}
\newcommand{\cD}{\mathcal{D}}
\newcommand{\cL}{\mathcal{L}}
\newcommand{\RR}{\mathbb{R}}

\newcommand{\EE}{\mathbb{E}}
\newcommand{\cH}{\mathcal{H}}

\def\ie{\emph{i.e.}, }

\newcommand{\ignore}[1]{}

\begin{document}
\pagestyle{headings}
\mainmatter

\title{Curriculum Manager for Source Selection\\
in Multi-Source Domain Adaptation\\
} 

\titlerunning{Curriculum Manager for Source Selection} 
\authorrunning{Yang et. al}
\author{Luyu Yang$^1$, Yogesh Balaji$^1$, Ser-Nam Lim$^2$, Abhinav Shrivastava$^1$}
\institute{$^1$University of Maryland \quad $^2$Facebook AI}

\maketitle

\begin{abstract}
The performance of Multi-Source Unsupervised Domain Adaptation depends significantly on the effectiveness of transfer from labeled source domain samples. In this paper, we proposed an adversarial agent that learns a dynamic curriculum for source samples, called Curriculum Manager for Source Selection (CMSS). The Curriculum Manager, an independent network module, constantly updates the curriculum during training, and iteratively learns which domains or samples are best suited for aligning to the target. The intuition behind this is to force the Curriculum Manager to constantly re-measure the transferability of latent domains over time to adversarially raise the error rate of the domain discriminator. CMSS does not require any knowledge of the domain labels, yet it outperforms other methods on four well-known benchmarks by significant margins. We also provide interpretable results that shed light on the proposed method. 
\end{abstract}

\section{Introduction}

Training deep neural networks requires datasets with rich annotations that are often time-consuming to obtain. Previous proposals to mitigate this issue have ranged from unsupervised~\cite{lee2019drop, kang2019contrastive, you2019towards, ouyang2019data, chen2019progressive, pan2019transferrable}, self-supervised~\cite{sun2019unsupervised, jeong2019self, valada2019self, yoon2019self}, to low shot learning~\cite{motiian2017few, saito2018maximum, wang2019few, zhang2019few}. Unsupervised Domain Adaptation (UDA), when first introduced in~\cite{ganin2014unsupervised}, sheds precious insights on how adversarial training can be utilized to get around the problem of expensive manual annotations. UDA aims to preserve the performance on an unlabeled dataset (target) using a model trained on a label-rich dataset (source) by making optimal use of the learned representations from the source. 

Intuitively, one would expect that having more labeled samples in the source domain will be beneficial. However, having more labeled samples does not equal better transfer, since the source will inadvertently encompass a larger variety of domains. While the goal is to learn a common representation for both source and target in such a Multi-Source Unsupervised Domain Adaptation (MS-UDA) setting, enforcing each source domain distribution to exactly match the target may increase the training difficulty, and generate ambiguous representations near the decision boundary potentially resulting in negative transfer. Moreover, for practical purposes, we would expect the data source to be largely unconstrained, whereby neither the number of domains or domain labels are known. A good example here would be datasets collected from the Internet where images come from unknown but potentially a massive set of users.
 
To address the MS-UDA problem, we propose an adversarial agent that learns a dynamic curriculum~\cite{weston2009curriculum} for multiple source domains, named Curriculum Manager for Source Selection (CMSS). More specifically, a constantly updated curriculum during training learns which domains or samples are best suited for aligning to the target distribution. The CMSS is an independent module from the feature network and is trained by maximizing the error of discriminator in order to weigh the gradient reversal back to the feature network. In our proposed adversarial interplay with the discriminator, the Curriculum Manager is forced to constantly re-measure the transferability of latent domains across time to achieve a higher error of the discriminator. Such a procedure of weighing the source data is modulated over the entire training. In effect, the latent domains with different transferability to the target distribution will gradually converge to different levels of importance without any need for additional domain partitioning prior or clustering.

We attribute the following contributions to this work:
\begin{itemize}
    \item We propose a novel adversarial method during training towards the MS-UDA problem. Our method does not assume any knowledge of the domain labels or the number of domains.
    \item Our method achieves state-of-the-art in extensive experiments conducted on four well-known benchmarks, including the large-scale DomainNet ($\sim$ 0.6 million images).
    \item We obtain interpretable results that show how CMSS is in effect a form of curriculum learning that has great effect on MS-UDA when compared to the prior art. This positively differentiates our approach from previous state-of-the-art.
\end{itemize}

\begin{figure*}[t]
\centering
    \includegraphics[width=1.0\textwidth]{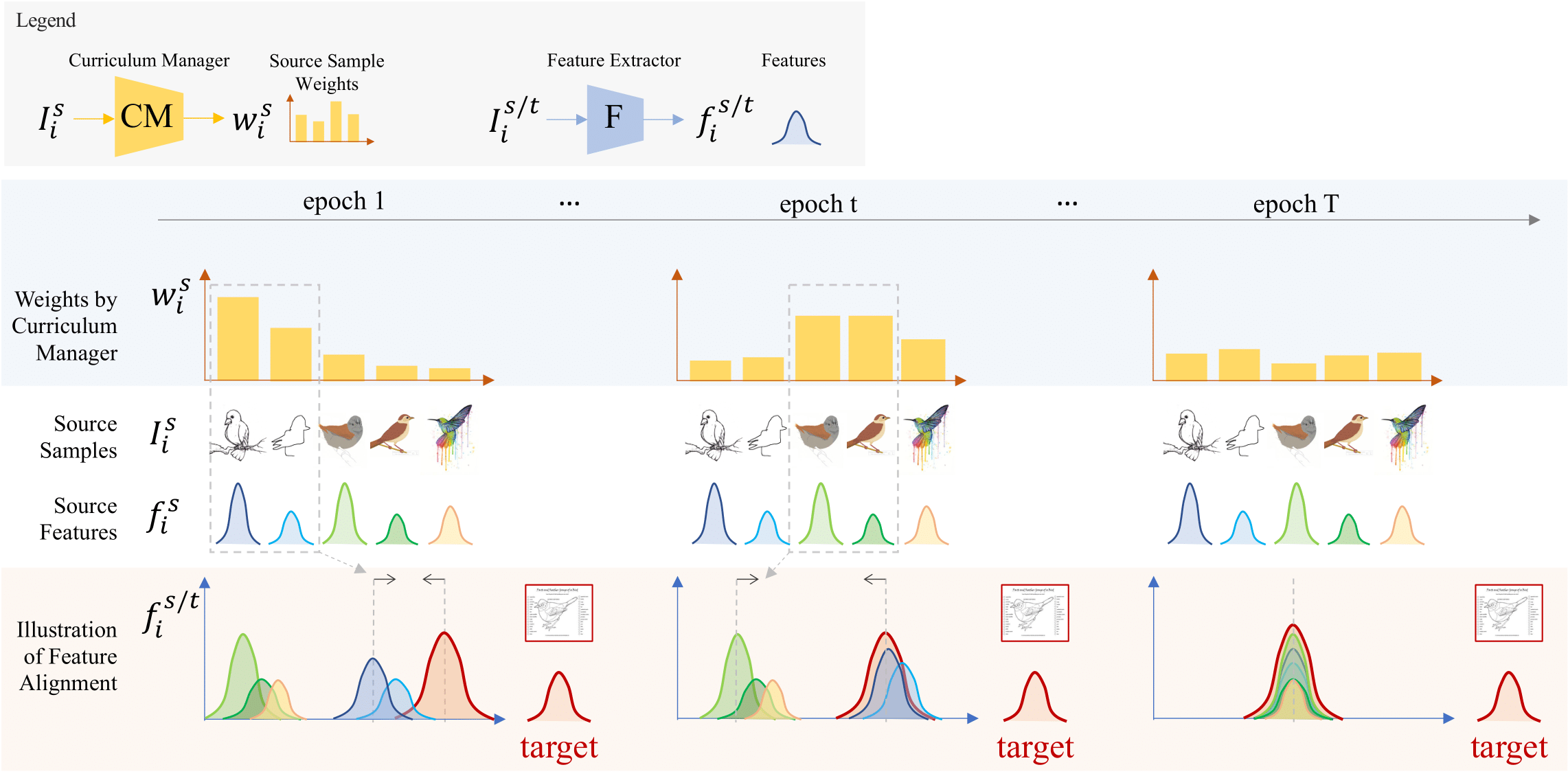}
    \caption{Illustration of CMSS during training. All training samples are passed through the feature network $F$. CMSS prefers samples with better transferability to match the target, and re-measure the transferability at each iteration to keep up with the discriminator. At the end of training after the majority of samples are aligned, the CMSS weights tend to be similar among source samples.}
    \label{fig:teaser}
\end{figure*}

\section{Related Work}

UDA is an actively studied area of research in machine learning and computer vision. Since the seminal contribution of Ben-David \textit{et al.} \cite{ben2007analysis,ben2010theory}, several techniques have been proposed for learning representations invariant to domain shift~\cite{li2018deep, ding2018semi,luo2019taking, chen2018domain,zhao2019learning}. In this section, we review some recent methods that are most related to our work.

\begin{figure*}[t]
\centering
    \includegraphics[width=1.05\textwidth]{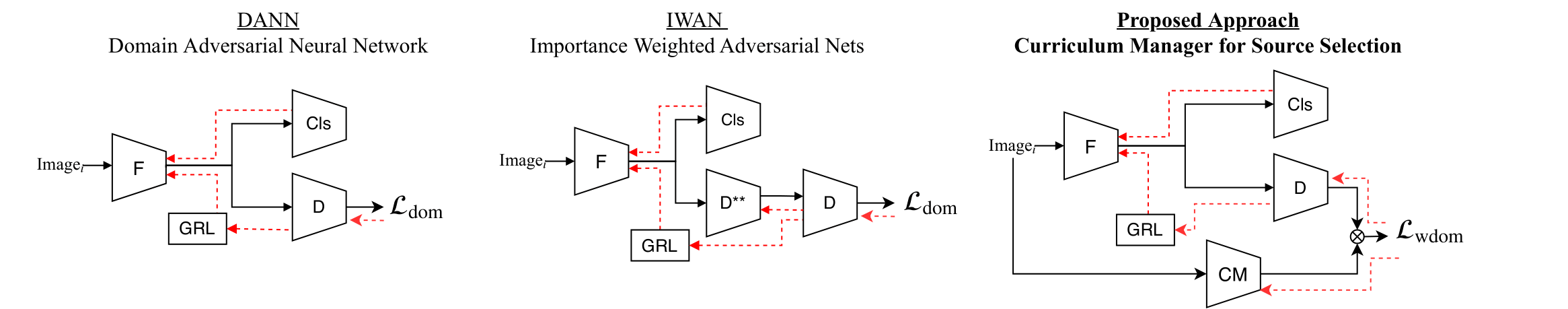}
    \caption{Architecture comparison of \textit{left}: DANN \cite{ganin2014unsupervised}, \textit{middle}: IWAN \cite{zhang2018importance}, and \textit{right}: proposed method. Red dotted lines indicate backward passes. ($F$: feature extractor, $Cls$: classifier, $D$: domain discriminator, $\text{GRL}$: gradient reversal layer, $\text{CM}$: Curriculum Manager, $\cL_{\text{dom}}$: Eq.\ref{eq:opt} domain loss, $\cL_{\text{wdom}}$: Eq.\ref{eq:dom} weighted domain loss)}
    \label{fig:network_fig}
\end{figure*}

\textbf{Multi-Source Unsupervised Domain Adaptation} (MS-UDA) assumes that the source training examples are inherently multi-modal. The source domains contain labeled samples while the target domain contains unlabeled samples~\cite{li2017deeper,peng2018moment,mancini2018boosting,ganin2014unsupervised,zhao2018adversarial}. In \cite{peng2018moment}, adaptation was performed by aligning the moments of feature distributions between each source-target pair. Deep Cocktail Network (DCTN) \cite{xu2018deep} considered the more realistic case of existence of category shift in addition to the domain shift, and proposes a $k$-way domain adversarial classifier and category classifier to generate a combined representation for the target.

Because domain labels are hard to obtain in the real world \ignore{multi-modal }datasets, latent domain discovery \cite{mancini2018boosting} -- a technique for alleviating the need for explicit domain label annotation has many practical applications. Xiong \textit{et al.} \cite{xiong2014latent} proposed to use square-loss mutual information based clustering with category distribution prior\ignore{ in each domain} to infer the domain assignment for images. Mancini \textit{et al.} \cite{mancini2018boosting} used a domain prediction branch to guide domain discovery using  multiple batch-norm layers.

\textbf{Domain-Adversarial Training} has been widely used \cite{chen2018re,cao2018partial,pei2018multi} since Domain-Adversarial Neural Network (DANN) \cite{ganin2014unsupervised} was proposed. The core idea is to train a discriminator network to discriminate source features from target, and train the feature network to fool the discriminator. Zhao \textit{et al.} \cite{zhao2018adversarial} first proposed to generalize DANN to the multi-source setting, and provides theoretical insights on the multi-domain adversarial bounds.
Maximum Classifier Discrepancy (MCD) \cite{saito2018maximum} is another powerful \cite{peng2018moment,kumar2018co,wu2019ace,liu2019transferable} technique for performing adaptation in an adversarial manner using two classifiers. The method first updates the classifiers to maximize the discrepancy between the classifiers' prediction on target samples, followed by minimizing the discrepancy while updating the feature generator.

\textbf{Domain Selection and Weighting:} 
Some previous methods that employed sample selection and sample weighing techniques for domain adaptation include \cite{duan2012exploiting,duan2012domain,duan2009domain}. Duan \textit{et al.} \cite{duan2012domain} proposed using a domain selection machine by leveraging a large number of loosely labeled web images from different sources. The authors of \cite{duan2012domain} adopted a set of base classifiers to predict labels for the target domain as well as a domain-dependent regularizer based on smoothness assumption. Bhatt \textit{et al.} \cite{bhatt2016multi} proposed to adapt iteratively by selecting the best sources that learn shared representations faster. Chen \textit{et al.} \cite{chen2018re} used a hand-crafted re-weighting vector so that the source domain label distribution is similar to the unknown target label distribution. Mancini \textit{et al.} \cite{mancini2019adagraph} modeled the domain dependency using a graph and utilizes auxiliary metadata for predictive domain adaptation. Zhang \textit{et al.} \cite{zhang2018importance} employed an extra domain classifier that gives the probability of a sample coming from the source domain. The higher the confidence is from such an extra classifier, the more likely it can be discriminated from the target domain, in which case the importance of the said sample is reduced accordingly.

\textbf{Curriculum for Domain Adaptation} aims at an adaptive strategy over time in order to improve the effectiveness of domain transfer. The curriculum can be hand-crafted or learned. Shu \textit{et. al} \cite{shu2019transferable} designed the curriculum by combining the classification loss and discriminator's loss as a weighting strategy to eliminate the corrupted samples in the source domain. Another work with similar motivation is \cite{chen2019progressive}, in which Chen \textit{et. al} proposed to use per-category prototype to measure the prediction confidence of target samples. A manually designed threshold $\tau$ is utilized to make a binary decision in selecting partial target samples for further alignment. Kurmi \textit{et. al} \cite{kurmi2019curriculum} used a curriculum-based dropout discriminator to simulate the gradual increase of sample variance.


\section{Preliminaries}
\label{sec:preliminaries}

\subsubsection{Task Formulation:}
\label{sec:setup}
In multi-source unsupervised domain adaptation (MS-UDA), we are given an input dataset $\cD_\text{src} = \{ (\bx_{i}^{s}, y_{i}^{s}) \}_{i=1}^{N_s}$ that contains samples from multiple domains. In this paper, we focus on classification problems, with the set of labels $y_{i}^{s} \in \{ 1, 2, \hdots, n_{c} \}$, where $n_c$ is the number of classes. Each sample $\bx_{i}^{s}$ has an associated domain label, $d_{i}^{s} \in \{ 1, 2, \hdots, S \}$, where $S$ is the number of source domains. In this work, we assume source domain label information is not known \textit{a priori}, i.e., number of source domains or source domain label per sample is not known.
In addition, given an unlabeled target dataset $\cD_\text{tgt} = \{ \bx_{i}^{t} \}_{i=1}^{N_t}$, the goal of MS-UDA is to train models using multiple source domains ($\cD_\text{src}$) and the target domain ($\cD_\text{tgt}$), and improve performance on the target test set.

\subsubsection{Domain-Adversarial training:}
\label{multi-src dann}
First, we discuss the domain-adversarial training formulation from \cite{ganin2014unsupervised} that is the basis from which we extend to MS-UDA. 
The core idea of domain-adversarial training is to minimize the distributional distance between source and target feature distributions posed as an adversarial game. The model has a feature extractor, a classifier, and a domain discriminator. 
The classifier takes in feature from the feature extractor and classifies it in $n_c$ classes. The discriminator is optimized to discriminate source features from target. The feature network, on the other hand, is trained to fool the discriminator while at the same time achieve good classification accuracy. 

More formally, let $F_{\theta}: \RR^{3 \times w \times h} \to \RR^{d}$ denote the feature extraction network, $C_{\phi}: \RR^{d} \to \RR^{n_c}$ denote the classifier, and $D_{\psi}: \RR^{d} \to \RR^{1}$ denote the domain discriminator. Here, $\theta$, $\phi$ and $\psi$ are the parameters associated with the feature extractor, classifier, and domain discriminator respectively. 
The model is trained using the following objective function:
\begin{align}\label{eq:opt}
    &\max_{\psi}~\min_{\theta, \phi}~~ \cL_\text{cls} - \lambda \cL_\text{dom} \\
  \text{where }~~~~ \cL_\text{cls} =& -\frac{1}{N_{s}} \sum_{i=1}^{N_{s}} \tby_{i} \log(C(F(\bx^{s}_{i}))) \nonumber \\
    \cL_\text{dom} =& -\EE_{\bx \sim \cD_\text{src}} \log(D(F(\bx))) - \EE_{\bx \sim \cD_\text{tgt}} \log(1 - D(F(\bx))) \nonumber \\
     =& -\frac{1}{N_{s}} \sum_{i=1}^{N_{s}} \log(D(F(\bx^{s}_{i}))) - \frac{1}{N_{t}} \sum_{i=1}^{N_{t}} \log(1 - D(F(\bx^{t}_{i}))) \nonumber
\end{align}
$\cL_\text{cls}$ is is the cross-entropy loss in source domain (with $\tby_{i}$ being the one-hot encoding of the label $y_i$), and $\cL_\text{dom}$ is the discriminator loss that discriminates source samples from the target. Note that both these loss functions use samples from all source domains.

In principle, if domain labels are available, there are two possible choices for the domain discriminator: (1) $k$ domain discriminators can be trained, each one discriminating one of the source domains from the target~\cite{ganin2014unsupervised}, or (2) a domain discriminator can be trained as a $(k+1)$-way classifier to classify input samples as either one of the source domains or target~\cite{zhao2018adversarial}. However, in our setup, domain labels are unknown and, therefore, these formulations can not be used.

\section{CMSS: Curriculum Manager for Source Selection}
\label{Sec3:our method}

For the source domain that is inherently multi-modal, our goal is to learn a dynamic curriculum for selecting the best-suited samples for aligning to the target feature distribution. At the beginning of training, the Curriculum Manager is expected to prefer samples with higher \emph{transferability} for aligning with the target, \ie source samples which have similar feature distributions to the target sample. Once the feature distributions of these samples are aligned, our Curriculum Manager is expected to prioritize the next round of source samples for alignment. As the training progresses, the Curriculum Manager can learn to focus on different aspects of the feature distribution as a proxy for better transferability. Since our approach learns a curriculum to prefer samples from different source domains, we refer to it is Curriculum Manager for Source Selection (CMSS).

Our approach builds on the domain-adversarial training framework (described in $\S$\ref{sec:preliminaries}). 
In this framework, our hypothesis is that source samples that are hard for the domain discriminator to separate from the target samples are likely the ones that have similar feature distributions. Our CMSS leverages this and uses the discriminator loss to find source samples that should be aligned first. The preference for source samples is represented as per-sample weights predicted by CMSS. Since our approach is based on domain-adversarial training, weighing $\cL_\text{dom}$ using these weights will lead to the discriminator encouraging the feature network to bring the distributions of higher weighted source samples closer to the target samples. This signal between the discriminator and feature extractor is achieved using the gradient reversal layer (see \cite{ganin2014unsupervised} for details).

Therefore, our proposed CMSS is trained to predict weights for source samples at each iteration, which maximizes the error of the domain discriminator. Due to this adversarial interplay with the discriminator, the CMSS is forced to re-estimate the preference of source samples across training to keep up with the improving domain discriminator. The feature extractor, $F$, is optimized to learn features that are both good for classification and confuse the discriminator. To avoid any influence from the classification task in the curriculum design, our CMSS also has an independent feature extractor module that learns to predict weights per-sample given the source images and domain discriminator loss.

\subsubsection{Training CMSS:} 
The CMSS weight for every sample in the source domain, $\bx^{s}_{i}$, is given by $w^{s}_{i}$. We represent this weighted distribution as $\tilde{\cD}_\text{src}$. The CMSS network is represented by $G_{\rho}: \RR^{c \times w \times h} \to \RR^{1}$ with parameters $\rho$. Given a batch of samples, $\bx^{s}_{1}, \bx^{s}_{2}, \hdots \bx^{s}_{b}$, we first pass these samples to $G_{\rho}$ to obtain an array of scores that are normalized using softmax function to obtain the resulting weight vector.
During training, the CMSS optimization objective can be written as
\begin{align}\label{eq:weight}
    \min_{\rho} \left[\frac{1}{N_{s}} \sum_{i=1}^{N_{s}} G_{\rho}(\bx^s_{i}) \log(D(F(\bx^{s}_{i})))\right]
\end{align}

With the source sample weights generated by CMSS, the loss function for domain discriminator can be written as
\begin{align}\label{eq:dom}
    \cL_\text{wdom} &= -\frac{1}{N_{s}} \sum_{i=1}^{N_{s}} G_{\rho}(\bx^s_{i}) \log(D(F(\bx^{s}_{i}))) -\frac{1}{N_{t}} \sum_{i=1}^{N_{t}} \log(1 - D(F(\bx^{t}_{i}))) \nonumber \\
                & ~~~~\text{ s.t. } \sum_{i} G_{\rho}(\bx^s_{i}) = N_s
\end{align}
The overall optimization objective can be written as
\begin{align}\label{eq:overall_objective}
    \max_{\psi}~ \min_{\theta, \phi, \rho} ~~& \cL_\text{cls} - \lambda \cL_\text{wdom}
\end{align}
where $\cL_\text{cls}$ is the Cross-Entropy loss for source classification and $\cL_\text{wdom}$ is the weighted domain discriminator loss from Eq.~\eqref{eq:dom}, with weights obtained by optimizing Eq.~\eqref{eq:weight}. $\lambda$ is the hyperparameter in the gradient reversal layer. We follow~\cite{ganin2014unsupervised} and set $\lambda$ based on the following annealing schedule: $\lambda_p = \frac{2}{1 + \exp(-\gamma \cdot p)} - 1$,
where $p$ is the current number of iterations divided by the total. $\gamma$ is set to $10$ in all experiments as in~\cite{ganin2014unsupervised}. Details of training are provided in Algorithm \ \ref{alg:proposed}.

\begin{algorithm}[b]
\caption{Training CMSS (Curriculum Manager for Source Selection)}
\label{alg:proposed}
\begin{algorithmic}[1]
\Require $N_\text{iter}$: Total number of training iterations
\Require $\gamma$: For computing $\lambda_p$ for $\cL_\text{wdom}$
\Require $N_{b}^{s}$ and $N_{b}^{t}$: Batch size for source and target domains
\State Shuffle the source domain samples
\For{$t$ in $(1:N_\text{iter})$}
    \State Compute $\lambda$ according to $2/(1 + \exp(-\gamma \cdot (t/N_{iter}))) - 1$
    \State Sample a training batch from source domains $\{ (\bx_i^s, y_i) \}_{i=1}^{N_{b}^s} \sim \cD_\text{src}$ and from target domain $\{ \bx_i^t \}_{i=1}^{N_{b}^t} \sim \cD_\text{tgt}$
    \State Update $\rho$ by $\min_{\rho}- \lambda  \cL_\text{wdom}$
    \State Update $\psi$ by $\min_{\psi}\lambda \cL_\text{dom}$
    \State Update $\theta, \phi$ by $\min_{\theta, \phi} \cL_\text{cls} - \lambda  \cL_\text{wdom}$

\EndFor 
\end{algorithmic}
\end{algorithm}

\subsection{CMSS: Theoretical Insights}
We first state the classic generalization bound for domain adaptation~\cite{NIPS2006_2983, NIPS2007_3212}. Let $\cH$ be a hypothesis space of $VC$-dimension $d$. For a given hypothsis class $\cH$, define the symmetric difference operator as $\cH \Delta \cH = \{ h(\bx) \oplus h'(\bx) | h, h' \in \cH \}$. Let $\cD_\text{src}$, $\cD_\text{tgt}$ denote the source and target distributions respectively, and $\hat{\cD}_\text{src}$, $\hat{\cD}_\text{tgt}$ denote the empirical distribution induced by sample of size $m$ drawn from $\cD_\text{src}$, $\cD_\text{tgt}$ respectively. Let $\epsilon_{s}$ ($\epsilon_{t}$) denote the true risk on source (target) domain, and $\hat{\epsilon}_{s}$ ($\hat{\epsilon}_{t}$) denote the empirical risk on source (target) domain. Then, following Theorem 1 of \cite{NIPS2007_3212}, with probability of at least $1-\delta$, $\forall h \in \cH$ ,
\begin{align}\label{eq:bound_orig}
    \epsilon_{t}(h) \leq \hat{\epsilon}_{s}(h) + \frac{1}{2} d_{\cH \Delta \cH}(\hat{\cD}_{src}, \hat{\cD}_{tgt}) + C
\end{align}
where $C$ is a constant 
\begin{align*}
    C = \lambda + O \left( \sqrt{ \frac{d\log(m/d) + \log(1/\delta)}{m} } \right)
\end{align*}
Here, $\lambda$ is the optimal combined risk (source + target risk) that can be achieved by hypothesis in $\cH$. Let 
$\{ \bx^{s}_{i} \}_{i=1}^{m}$, $\{ \bx^{t}_{i} \}_{i=1}^{m}$ be the samples in the empirical distributions $\hat{\cD}_\text{src}$ and $\hat{\cD}_\text{tgt}$ respectively. Then, $P(\bx^{s}_{i}) = 1/m$ and $P(\bx^{t}_{i}) = 1/m$. The empirical source risk can be written as $\hat{\epsilon}_{s}(h) = 1/m \sum_{i}\hat{\epsilon}_{\bx^{s}_{i}}(h)$

Now consider a CMSS re-weighted source distribution $\hat{\cD}_\text{wsrc}$, with $P(\bx^{s}_{i}) = w_{i}$. For $\hat{\cD}_\text{wsrc}$ to be a valid probability mass function, $\sum_{i}w^{s}_{i} = 1$ and $w^{s}_{i} \geq 0$. Note that $\hat{\cD}_\text{src}$ and $\hat{\cD}_\text{wsrc}$ share the same samples, and only differ in weights. The generalization bound for this re-weighted distribution can be written as
\begin{align*}
    \epsilon_{t}(h) \leq \sum_{i} w_{i} \hat{\epsilon}_{\bx^{s}_{i}}(h) + \frac{1}{2} d_{\cH \Delta \cH}(\hat{\cD}_\text{wsrc}, \hat{\cD}_\text{tgt}) + C
\end{align*}
Since the bound holds for all weight arrays $\bw = [w^{s}_{1}, w^{s}_{2}\hdots w^{s}_{m}]$ in a simplex, we can minimize the objective over $\bw$ to get a tighter bound. 
\begin{align}\label{eq:weighted_bound}
    \epsilon_{t}(h) \leq \min_{\bw \in \Delta^{m}} \sum_{i} w_{i} \hat{\epsilon}_{\bx^{s}_{i}}(h) + \frac{1}{2} d_{\cH \Delta \cH}(\hat{\cD}_\text{wsrc}, \hat{\cD}_{\text{tgt}}) + C
\end{align}
The first term is the weighted risk, and the second term $d_{\cH \Delta \cH}(\hat{\cD}_\text{wsrc}, \hat{\cD}_\text{tgt})$ is the weighted symmetric divergence which can be realized using our weighted adversarial loss. Note that when $\bw = [1/m, 1/m, \hdots 1/m]$, we get the original bound~\eqref{eq:bound_orig}. Hence, the original bound is in the feasible set of this optimization.

\subsubsection{Relaxations.}
In practice, deep neural networks are used to optimize the bounds presented above. Since the bound~\eqref{eq:weighted_bound} is minimized over the weight vector $\bw$, one trivial solution is to assign non-zero weights to only a few source samples. In this case, a neural network can overfit to these source samples, which could result in low training risk and low domain divergence. To avoid this trivial case, we present two relaxations:
\begin{itemize}
    \item We use the unweighted loss for the source risk (first term in the bound~\eqref{eq:weighted_bound}).  
    \item For the divergence term, instead of minimizing $\bw$ over all the samples, we optimize only over mini-batches. Hence, for every mini-batch, there is at least one $w_{i}$ which is non-zero. Additionally, we make weights a function of input, \ie $w_{i} = G_{\rho}(\bx^{s}_{i})$, which is realized using a neural network. This will smooth the predictions of $w_{i}$, and make the weight network produce a soft-selection over source samples based on correlation with the target. 
\end{itemize}
Note that the $G_{\rho}$ network discussed in the previous section satisfies these criteria.


\section{Experimental Results}
\label{sec:experiment}
In this section, we perform an extensive evaluation of the proposed method on the following tasks: digit classification(\textit{MNIST, MNIST-M, SVHN, Synthetic Digits, USPS}), image recognition on the large-scale DomainNet dataset (\textit{clipart, infograph, paiting, quickdraw, real, sketch}), PACS\cite{li2017deeper} (\textit{art, cartoon, photo} and \textit{sketch}) and Office-Caltech10 (\textit{Amazon, Caltech, Dslr, Webcam}). We compare our method with the following contemporary approaches:
Domain Adversarial Neural Network (\textbf{DANN}) \cite{ganin2014unsupervised}, Multi-Domain Adversarial Neural Network (\textbf{MDAN})\cite{zhao2018adversarial} and two state-of-the-art discrepancy-based approaches: Maximum Classifier Discrepancy (\textbf{MCD}) \cite{saito2018maximum} and Moment Matching for Multi-Source (\textbf{$M^3$SDA}) \cite{peng2018moment}.
We follow the protocol used in other multi-source domain adaptation works \cite{mancini2018boosting, peng2018moment}, where each domain is selected as the target domain while the rest of domains are used as source domains.
For \textbf{Source Only} and \textbf{DANN} experiments, all source domains are shuffled and treated as one domain. To guarantee fairness of comparison, we used the same model architectures, batch size and data pre-processing routines for all compared approaches. All our experiments are implemented in PyTorch.

\subsection{Experiments on Digit Recognition}
Following DCTN~\cite{xu2018deep} and $M^3$SDA~\cite{peng2018moment}, we sample $25000$ images from \ignore{each }training subset and $9000$ from testing subset of \textit{MNIST, MNIST-M, SVHN} and \textit{Synthetic Digits}. The entire \textit{USPS} is used \ignore{as a domain }since it contains only $9298$ images in total.

In all the experiments, the feature extractor is composed of three $conv$ layers and two $fc$ layers. The entire network is trained from scratch with batch size equals $16$. For each experiment, we run the same setting five times and report the mean and standard deviation. (See \textit{Appendix} for more experiment details and analyses.)
The results are shown in Table \ref{tab:digits result}. The proposed method achieves an $\textbf{90.8\%}$ average accuracy, outperforming other baselines by a large margin ($\sim 3 \%$ improvement on the previous state-of-the-art approach). 

\begin{table}[t]
\centering
\renewcommand{\arraystretch}{1.0}
\renewcommand{\tabcolsep}{1.6mm}
 \caption{\textbf{Results on Digits classification}. The proposed CMSS achieves $\textbf{90.8\%}$ accuracy. Comparisons with MCD and $M^3SDA$ are reprinted from \cite{peng2018moment}. All experiments are based on a 3-$conv$-layer backbone trained from scratch. (\textcolor{blue}{mt, mm, sv, sy, up}: \textit{MNIST, MNIST-M, SVHN, Synthetic Digits, UPSP})}
\resizebox{\textwidth}{!}{
    \begin{tabular}{@{}lcccccc @{}}
    \toprule
    \multirow{2}{*}{Models} & \textcolor{blue}{$mm,sv,sy,up$} & \textcolor{blue}{$mt,sv,sy,up$}
    & \textcolor{blue}{$mt,mm,sy,up$} & \textcolor{blue}{$mt,mm,sv,up$} & \textcolor{blue}{$mt,mm,sv,sy$} & \multirow{2}{*}{Avg} \\
    & $\to$ \textcolor{blue}{$mt$} & $\to$ \textcolor{blue}{$mm$}
    & $\to$ \textcolor{blue}{$sv$} & $\to$ \textcolor{blue}{$sy$} & $\to$ \textcolor{blue}{$up$} & \\
    \midrule
    Source Only & 92.3 $\pm$ 0.91 & 63.7 $\pm$ 0.83 & 71.5 $\pm$ 0.75 & 83.4 $\pm$ 0.79 & 90.7 $\pm$ 0.54 & 80.3 $\pm$ 0.76\\
    \midrule
    DANN\cite{ganin2014unsupervised} & 97.9 $\pm$ 0.83 & 70.8 $\pm$ 0.94 & 68.5 $\pm$ 0.85 & 87.3 $\pm$ 0.68 & 93.4 $\pm$ 0.79 & 83.6 $\pm$ 0.82  \\
    MDAN\cite{zhao2018adversarial}  & 97.2 $\pm$ 0.98 & \textbf{75.7} $\pm$ 0.83 & 82.2 $\pm$ 0.82 & 85.2 $\pm$ 0.58 & 93.3 $\pm$ 0.48 & 86.7 $\pm$ 0.74 \\ 
    MCD\cite{saito2018maximum} &96.2 $\pm$ 0.81 & 72.5 $\pm$ 0.67 & 78.8 $\pm$ 0.78 & 87.4 $\pm$ 0.65 & 95.3 $\pm$ 0.74 & 86.1 $\pm$ 0.64 \\
    $M^3$SDA\cite{peng2018moment} & 98.4 $\pm$ 0.68 & 72.8 $\pm$ 1.13 & 81.3 $\pm$ 0.86 & 89.5 $\pm$ 0.56 & 96.1 $\pm$ 0.81 & 87.6 $\pm$ 0.75 \\
    \midrule
    CMSS & \textbf{99.0} $\pm$ 0.08 & 75.3 $\pm$ 0.57 & \textbf{88.4} $\pm$ 0.54 & \textbf{93.7} $\pm$ 0.21 & \textbf{97.7} $\pm$ 0.13 & \textbf{90.8} $\pm$ 0.31 \\
    \bottomrule
    \end{tabular}
    }
    \label{tab:digits result}
\end{table}

\subsection{Experiments on DomainNet}
Next, we evaluate our method on \textbf{DomainNet}~\cite{peng2018moment} -- a large-scale benchmark dataset used for multi-domain adaptation. The DomainNet dataset contains samples from $6$ domains: \textit{Clipart}, \textit{Infograph}, \textit{Painting}, \textit{Quickdraw}, \textit{Real} and \textit{Sketch}. Each domain has $\textbf{345}$ categories, and the dataset has $\sim$ \textbf{$\textbf{0.6}$ million} images in total, which is the largest existing domain adaptation dataset.
We use ResNet-101 pretrained on ImageNet as the feature extractor for in all our experiments. For CMSS, we use a ResNet-18 pretrained on ImageNet. The batch size is fixed to $128$. We conduct experiments over $5$ random runs, and report mean and standard deviation over the 5 runs.

The results are shown in Table \ref{tab:visda_results}. CMSS achieves $\textbf{46.5\%}$ average accuracy, outperforming other baselines by a large margin. We also note that our approach achieves the best performance in each experimental setting. It is also worth mentioning that in the experiment when the target domain is \textit{Quickdraw  (\textcolor{blue}{q})}, our approach is the only one that outperforms Source Only baseline, while all other compared approaches result in negative transfer (lower performance than the source-only model). This is since \textit{quickdraw} has a significant domain shift compared to all other domains. This shows that our approach can effectively alleviate negative transfer even in such challenging set-up.

\begin{table*}[t]
\centering
\renewcommand{\arraystretch}{1.2}
\renewcommand{\tabcolsep}{1.8mm}
\caption{\textbf{Results on the DomainNet dataset}. CMSS achieves $46.5\%$ average accuracy. When the target domain is \textit{quickdraw} $\textcolor{blue}{q}$, CMSS is the only one that outperforms Source Only which indicates \textit{negative transfer} has been alleviated. \textit{Source Only *} is re-printed from \cite{peng2018moment}, \textit{Source Only} is our implemented results. All experiments are based on ResNet-101 pre-trained on ImageNet. (\textcolor{blue}{$c$}: \textit{clipart},
   \textcolor{blue}{$i$}: \textit{infograph},
   \textcolor{blue}{$p$}: \textit{painting},
   \textcolor{blue}{$q$}: \textit{quickdraw},
   \textcolor{blue}{$r$}: \textit{real},
   \textcolor{blue}{$s$}: \textit{sketch})
   }
\resizebox{\textwidth}{!}{
    \begin{tabular}{@{}lccccccc@{}}
    \toprule
    \multirow{2}{*}{Models} & \textcolor{blue}{$i,p,q$} & \textcolor{blue}{$c,p,q$} & \textcolor{blue}{$c,i,q$} & \textcolor{blue}{$c,i,p$} & \textcolor{blue}{$c,i,p$} &\textcolor{blue}{$c,i,p$} & \multirow{2}{*}{Avg} \\
    & \textcolor{blue}{$r,s$} $\to$ \textcolor{blue}{$c$}
    & \textcolor{blue}{$r,s$} $\to$ \textcolor{blue}{$i$}
    & \textcolor{blue}{$r,s$} $\to$ \textcolor{blue}{$p$}
    & \textcolor{blue}{$r,s$} $\to$ \textcolor{blue}{$q$}
    & \textcolor{blue}{$q,s$} $\to$ \textcolor{blue}{$r$}
    & \textcolor{blue}{$q,r$} $\to$ \textcolor{blue}{$s$}
    &\\
    \midrule
    Source Only* & 47.6$\pm$0.52 & 13.0$\pm$0.41 & 38.1$\pm$0.45 & 13.3$\pm$0.39 & 51.9$\pm$0.85 & 33.7$\pm$0.54 & 32.9$\pm$0.54 \\
    Source Only & 52.1$\pm$0.51 & 23.4$\pm$0.28 & 47.7$\pm$0.96 & 13.0$\pm$0.72 & 60.7$\pm$0.32 & 46.5$\pm$0.56 & 40.6$\pm$0.56\\
    \midrule
    DANN\cite{ganin2014unsupervised} & 60.6$\pm$0.42 & 25.8$\pm$0.34 & 50.4$\pm$0.51 & 7.7$\pm$0.68 & 62.0$\pm$0.66 &  51.7$\pm$0.19 & 43.0$\pm$0.46 \\
    MDAN\cite{zhao2018adversarial}  & 60.3$\pm$0.41 & 25.0$\pm$0.43 & 50.3$\pm$0.36 & 8.2$\pm$1.92 & 61.5$\pm$0.46 & 51.3$\pm$0.58 & 42.8$\pm$0.69 \\
    MCD\cite{saito2018maximum} & 54.3$\pm$0.64 & 22.1$\pm$0.70 & 45.7$\pm$0.63 & 7.6$\pm$0.49 & 58.4$\pm$0.65 & 43.5$\pm$0.57 & 38.5$\pm$0.61  \\
    $M^3$SDA\cite{peng2018moment} & 58.6$\pm$0.53 & 26.0$\pm$0.89 & 52.3$\pm$0.55 & 6.3$\pm$0.58 & 62.7$\pm$0.51 & 49.5$\pm$0.76 & 42.6$\pm$0.64 \\
    \midrule
    CMSS & \textbf{64.2}$\pm$0.18 & \textbf{28.0}$\pm$0.20 & \textbf{53.6}$\pm$0.39 & \textbf{16.0}$\pm$0.12 & \textbf{63.4}$\pm$0.21 & \textbf{53.8}$\pm$0.35 & \textbf{46.5}$\pm$0.24 \\
    \bottomrule
    \end{tabular}
    }
   
   \label{tab:visda_results}
\end{table*}

\subsection{Experiments on PACS}
PACS \cite{li2017deeper} is another popular benchmark for multi-source domain adaptation. It contains 4 domains: \textit{art, cartoon, photo} and \textit{sketch}. Images of 7 categories are collected for each domain. There are $9991$ images in total.
For all experiments\ignore{ on PACS}, we used ResNet-18 pretrained on ImageNet as the feature extractor following \cite{mancini2018boosting}. For the Curriculum Manager, we use the same architecture as the feature extractor. Batch size of $32$ is used. We conduct experiments over $5$ random runs, and report mean and standard deviation over the runs.
The results \ignore{on PACS} are shown in Table~\ref{tab:pacs result} (\textcolor{blue}{$a$}: \textit{art}, \textcolor{blue}{$c$}: \textit{cartoon}, \textcolor{blue}{$p$}: \textit{painting}, \textcolor{blue}{$s$}: \textit{sketch}.). CMSS achieves the state-of-the-art average accuracy of $\textbf{89.5\%}$. On the most challenging \textit{sketch} (\textcolor{blue}{s}) domain, we obtain \ignore{an} $\textbf{82.0\%}$, outperforming other baselines by a large margin.

\renewcommand{\tabcolsep}{4pt}
\begin{table}[t]
    \begin{minipage}[t]{0.54\linewidth}
        \centering
        \footnotesize
        \caption{Results on PACS}
        \renewcommand{\arraystretch}{1.2}
	    \renewcommand{\tabcolsep}{1.8mm}
        \resizebox{\textwidth}{!}{
         \begin{tabular}{@{}l ccccc@{}}
    \toprule
   Models & \textcolor{blue}{$c,p,s$} $\to$ \textcolor{blue}{$a$}
   & \textcolor{blue}{$a,p,s$} $\to$ \textcolor{blue}{$c$}
   & \textcolor{blue}{$a,c,s$} $\to$ \textcolor{blue}{$p$}
   & \textcolor{blue}{$a,c,p$} $\to$ \textcolor{blue}{$s$}
   & Avg \\
    \midrule
    Source Only & 74.9$\pm$0.88 & 72.1$\pm$0.75 & 94.5$\pm$0.58 & 64.7$\pm$1.53 & 76.6$\pm$0.93\\
    \midrule
    DANN\cite{ganin2014unsupervised} & 81.9$\pm$1.13 & 77.5$\pm$1.26 & 91.8$\pm$1.21 & 74.6$\pm$1.03 & 81.5$\pm$1.16   \\ 
    MDAN\cite{zhao2018adversarial}  & 79.1$\pm$0.36 & 76.0$\pm$0.73 & 91.4$\pm$0.85 & 72.0$\pm$0.80 & 79.6$\pm$0.69 \\
    WBN\cite{mancini2018boosting} & \textbf{89.9}$\pm$0.28 & 89.7$\pm$0.56 & \textbf{97.4}$\pm$0.84 & 58.0$\pm$1.51 & 83.8$\pm$0.80 \\
    MCD\cite{saito2018maximum} & 88.7$\pm$1.01 & 88.9$\pm$1.53 & 96.4$\pm$0.42 & 73.9$\pm$3.94 & 87.0$\pm$1.73 \\
    $M^3$SDA\cite{peng2018moment} & 89.3$\pm$0.42 & 89.9$\pm$1.00 & 97.3$\pm$0.31 & 76.7$\pm$2.86 & 88.3$\pm$1.15 \\
    
    \midrule
    CMSS & 88.6$\pm$0.36 & \textbf{90.4}$\pm$0.80 & 96.9$\pm$0.27 & \textbf{82.0}$\pm$0.59 & \textbf{89.5}$\pm$0.50 \\
    \bottomrule
    \end{tabular}
        }
        \label{tab:pacs result}
    \end{minipage}
    \hfill
    \begin{minipage}[t]{0.44\linewidth}
    \centering
    \footnotesize
    \vfill
    \caption{Results on Office-Caltech10}
    \renewcommand{\arraystretch}{1.16}
    \renewcommand{\tabcolsep}{1.8mm}
    \resizebox{\textwidth}{!}{
     \begin{tabular}{@{}lccccc@{}}
    \toprule
    \multirow{2}{*}{Models} & \textcolor{blue}{$A,C,D$} & \textcolor{blue}{$A,C,W$}
    & \textcolor{blue}{$A,D,W$} & \textcolor{blue}{$C,D,W$} & \multirow{2}{*}{Avg} \\
    & $\to$ \textcolor{blue}{$W$} & $\to$ \textcolor{blue}{$D$}
    & $\to$ \textcolor{blue}{$C$} & $\to$ \textcolor{blue}{$A$} &\\
    \midrule
    Source Only & 99.0 & 98.3 & 87.8 & 86.1  & 92.8 \\
    \midrule
    DANN\cite{ganin2014unsupervised} & 99.3 & 98.2 & 89.7 & 94.8 & 95.5 \\
    MDAN\cite{zhao2018adversarial}  & 98.9 & 98.6 & 91.8 & 95.4 & 96.1  \\ 
    MCD\cite{saito2018maximum} & 99.5 & 99.1 & 91.5 & 92.1 & 95.6  \\
    $M^3$SDA\cite{peng2018moment} & 99.5 & 99.2 & 92.2 & 94.5 & 96.4 \\
    \midrule
    CMSS & \textbf{99.6} & \textbf{99.3} & \textbf{93.7} & \textbf{96.0} & \textbf{97.2} \\
    \bottomrule
    \end{tabular}
    }
    \label{tab:office caltech result}
    \end{minipage}
    \vspace{-0.15in}
\end{table}

\subsection{Experiments on Office-Caltech10}
The office-Caltech10 \cite{gong2012geodesic} dataset has 10 object categories from 4 different domains: \textit{Amazon, Caltech, DSLR}, and \textit{Webcam}. For all the experiments, we use the same architecture (ResNet-101 pretrained on ImageNet) used in \cite{peng2018moment}. The experimental results are shown in Table \ref{tab:office caltech result} (\textcolor{blue}{A}: \textit{Amazon}, \textcolor{blue}{C}: \textit{Caltech}, \textcolor{blue}{D}: \textit{Dslr}, \textcolor{blue}{W}: \textit{Webcam}).  CMSS achieves state-of-the-art average accuracy of \textbf{97.2$\%$}.

\subsection{Comparison with other re-weighting methods}
\begin{figure}[t]
\centering
    \includegraphics[width=1.0\textwidth]{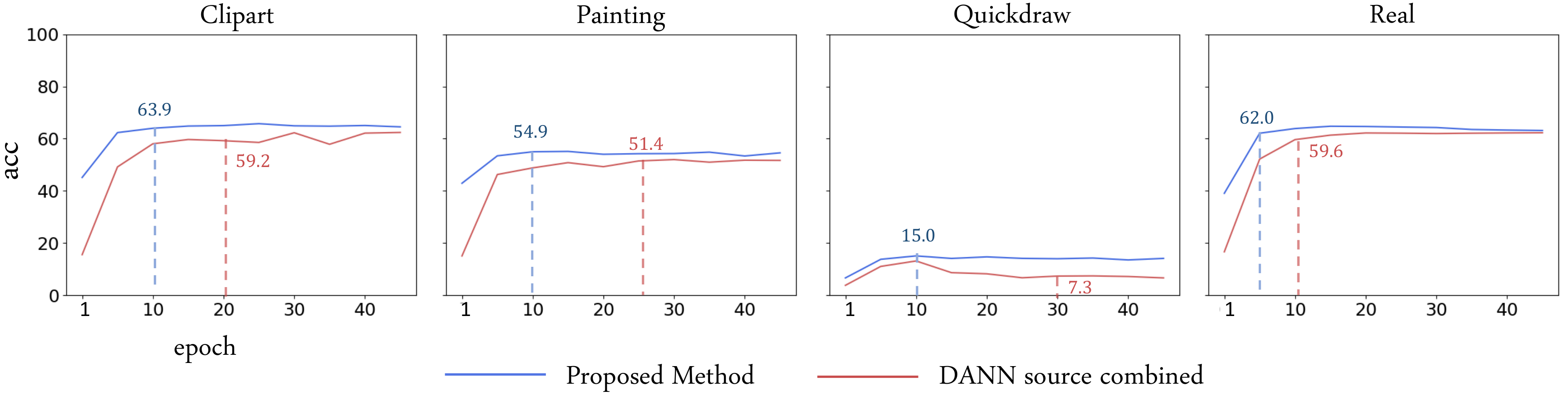}
    \caption{Test accuracy after the model is trained for $t$ epochs. Comparison between CMSS and DANN using source domains combined as one.}
    \label{fig:test_acc_over_time}
\end{figure}

\ignore{Besides network architecture comparison with IWAN \cite{zhang2018importance}, we compare the experiment results and the weighting changes over time between IWAN and the proposed CMSS. The comparison results on DomainNet are shown in Table \ref{tab:compare_iwan} (abbreviations of domains same as Table \ref{tab:visda_results}). IWAN obtained an $43.1\%$ average accuracy which is close to using DANN with combined source domains. We analyze the weighting changes over the training epochs and find that in IWAN the weighting for source samples are all close to 1. The comparison weight mean $\pm$ variance over time is shown in Figure \ref{fig:w_variance} with more discussion in section \ref{sec:selection_over_time}.

Moreover, CMSS also achieved a faster and more stable convergence with test accuracy compared with DANN \cite{ganin2014unsupervised} using combined source (Figure \ref{fig:test_acc_over_time}), which illustrates the effectivness of the learnt curriculum.
}

In this experiment, we compare CMSS with other weighing schemes proposed in the literature. We use IWAN~\cite{zhang2018importance} for this purpose. IWAN, originally proposed for partial domain adaption, reweights the samples in adversarial training using outputs of discriminator as sample weights (Refer to Figure~\ref{fig:network_fig}). CMSS, however, computes sample weights using a separate network $G_{\rho}$ updated using an adversarial game. We adapt IWAN for multi-source setup and compare it against our approach. The results are shown in Table \ref{tab:compare_iwan} (abbreviations of domains same as Table \ref{tab:visda_results}). IWAN obtained $43.1\%$ average accuracy which is close to performance obtained using DANN with combined source domains. For further analysis, we plot how sample weights estimated by both approaches (plotted as mean $\pm$ variance) change as training progresses in Figure \ref{fig:w_variance}. We observe that CMSS selects weights with larger variance which demonstrates its sample selection ability, while IWAN has weights all close to $1$ (in which case, it becomes similar to DANN). This illustrates the superiority of our sample selection method. More discussions on sample selection can be found in Section \ref{sec:selection_over_time}. CMSS also achieves a faster and more stable convergence in test accuracy compared to DANN \cite{ganin2014unsupervised} where we assume a single source domain (Figure \ref{fig:test_acc_over_time}), which further supports the effectiveness of the learnt curriculum.

\renewcommand{\tabcolsep}{4pt}
\begin{table}[t]
    \begin{minipage}[t]{0.53\linewidth}
        \centering
        \footnotesize
        \caption{Comparing re-weighting methods}
        \renewcommand{\arraystretch}{1.1}
	    \renewcommand{\tabcolsep}{0.8mm}
        \resizebox{\textwidth}{!}{
        \begin{tabular}{@{}l ccccccc@{}}
        \toprule
        \multirow{2}{*}{Models} & \textcolor{blue}{$i,p,q$} & \textcolor{blue}{$c,p,q$} & \textcolor{blue}{$c,i,q$} & \textcolor{blue}{$c,i,p$} & \textcolor{blue}{$c,i,p$} &\textcolor{blue}{$c,i,p$} & \multirow{2}{*}{Avg} \\
        & \textcolor{blue}{$r,s$} $\to$ \textcolor{blue}{$c$}
        & \textcolor{blue}{$r,s$} $\to$ \textcolor{blue}{$i$}
        & \textcolor{blue}{$r,s$} $\to$ \textcolor{blue}{$p$}
        & \textcolor{blue}{$r,s$} $\to$ \textcolor{blue}{$q$}
        & \textcolor{blue}{$q,s$} $\to$ \textcolor{blue}{$r$}
        & \textcolor{blue}{$q,r$} $\to$ \textcolor{blue}{$s$}
        &\\
        \midrule
        DANN\cite{ganin2014unsupervised} & 60.6 & 25.8 & 50.4 & 7.7 & 62.0 & 51.7 & 43.0 \\
        IWAN\cite{zhang2018importance} & 59.1 & 25.2 & 49.7 & 12.9 & 60.4 & 51.4 & 43.1\\
        \midrule
        CMSS & \textbf{64.2}& \textbf{28.0} & \textbf{53.6} & \textbf{16.0} & \textbf{63.4} & \textbf{53.8} & \textbf{46.5} \\
        \bottomrule
        \end{tabular}
        }
        \label{tab:compare_iwan}
    \end{minipage}
    \hfill
    \begin{minipage}[t]{0.45\linewidth}
    \centering
    \footnotesize
    \vfill
    \includegraphics[width=1.0\textwidth]{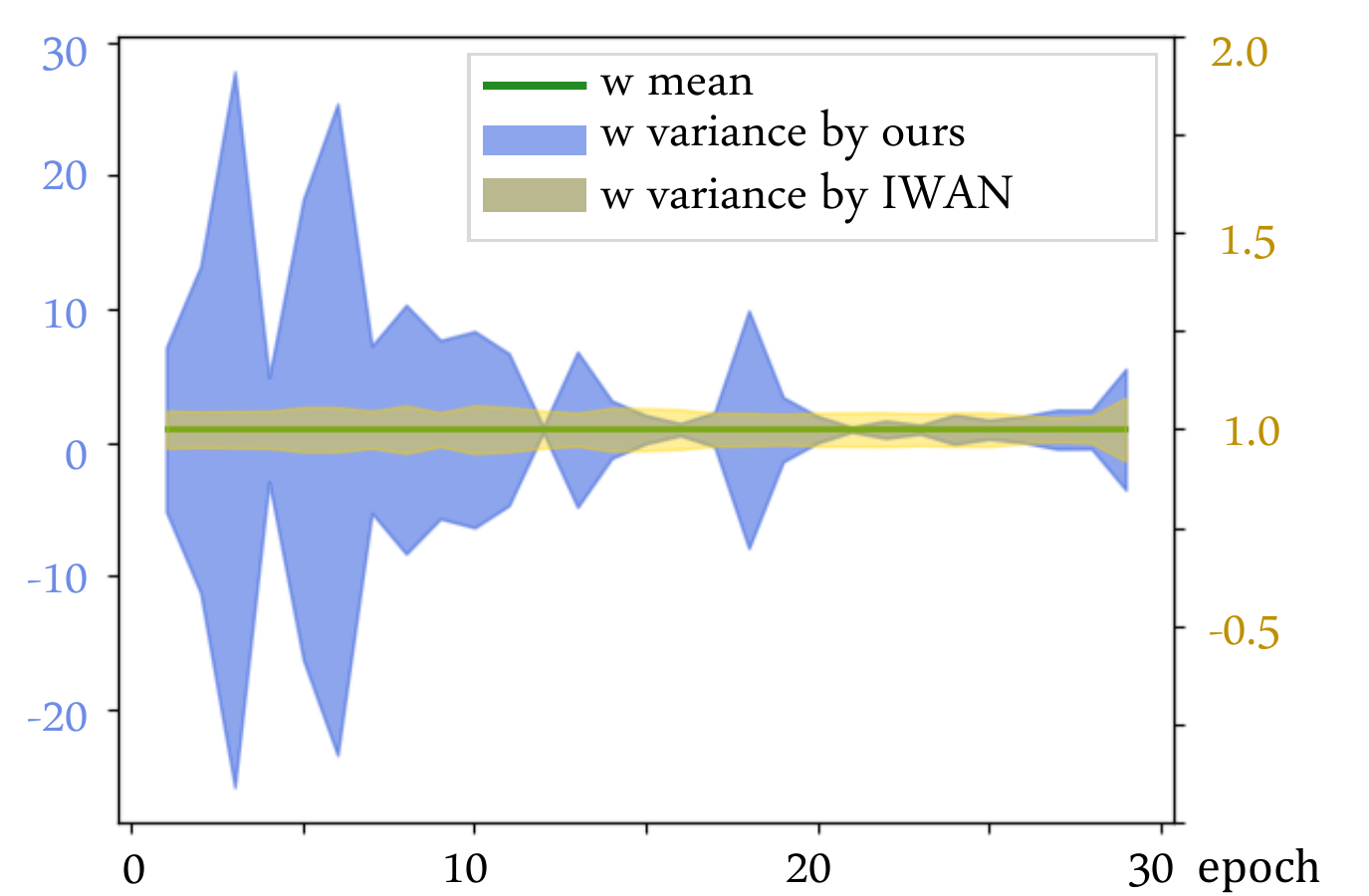}
    \captionof{figure}{Mean/var of weights over time.}
    \label{fig:w_variance}
    \end{minipage}
    \vspace{-0.15in}
\end{table}

\begin{figure*}[t]
\centering
    \includegraphics[width=0.95\textwidth]{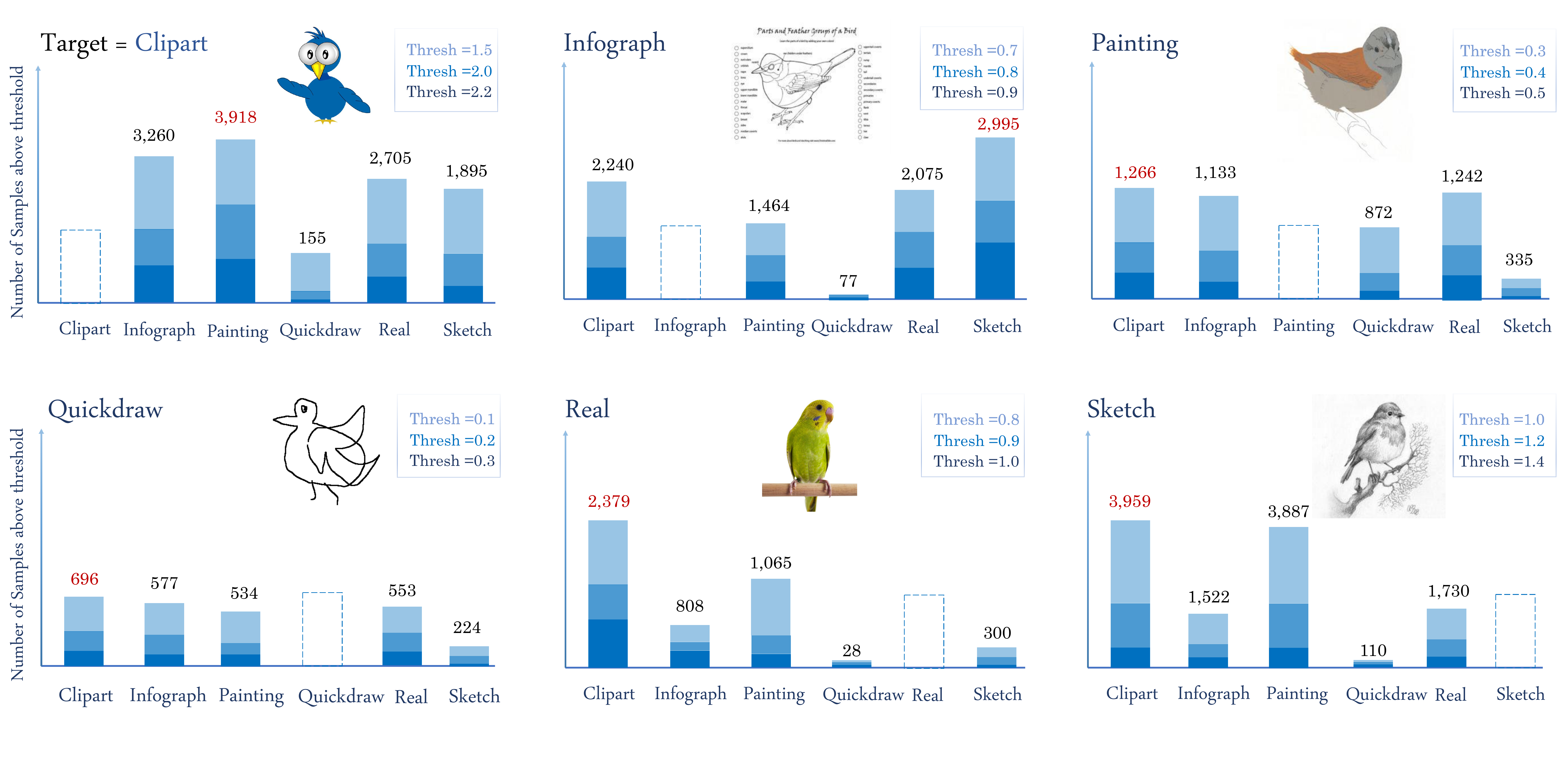}
    \caption{\textbf{Interpretation results of the sample selection} on DomainNet dataset using the proposed method. In each plot, one domain is selected as the target. In each setting, predictions of CMSS are computed for each sample of the source domains. The bars indicate how many of these samples have weight prediction larger than a manually chosen threshold, with each bar denoting a single source domain. Maximum number of samples are highlighted in red. \textit{Best viewed in color}}
    \label{fig:w_selection_all}
\end{figure*}

\begin{figure}[t]
\centering
    \includegraphics[width=1.00\textwidth]{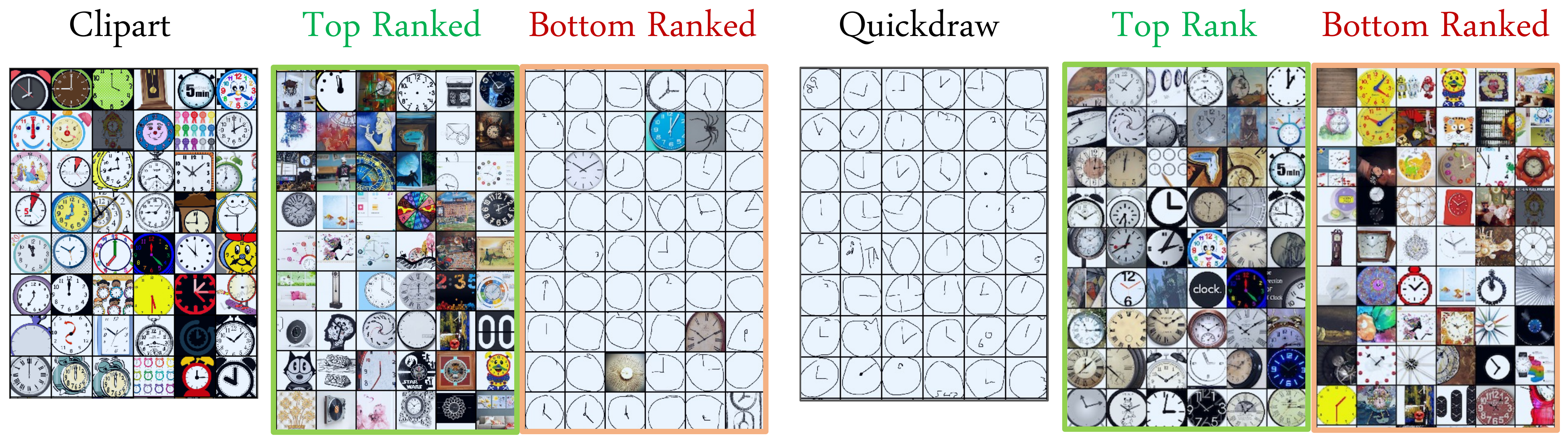}
    \caption{Ranked source samples according to learnt weights (class ``Clock'' of DomainNet dataset). \textit{LHS}: Examples of unlabeled target domain \textit{Clipart} and the Top/Bottom Ranked $\sim$ 50 samples of the source domain composed of \textit{Infograph, Painting, Quickdraw, Real} and \textit{Sketch}. \textit{RHS}: Examples of unlabeled target domain \textit{Quickdraw} and the Ranked samples of source domain composed of \textit{Clipart, Infograph, Painting, Real} and \textit{Sketch}. Weights are obtained at inference time using CMSS trained after $5$ epochs.}
    \label{fig:w_ranked_clock}
\end{figure}


\section{Interpretations}
In this section, we are interested in understanding and visualizing the source selection ability of our approach. We conduct two sets of experiments: (i) visualizations of the source selection curriculum over time, and (ii) comparison of our selection mechanism with other sample re-weighting methods.

\subsection{Visualizations of source selection}

\subsubsection{Domain Preference} 
\ignore{As previously argued in MDAN \cite{zhao2018adversarial} and WBN \cite{mancini2018boosting}, weighting the source domains (or latent source domains) according to their similarity to the target domain would alleviate the domain shift and results in improved domain alignment for multi-source DA tasks. }
We first investigate if CMSS indeed exhibits domain preference over the course of training as claimed. For this experiment, we randomly select $m=34000$ training samples from each source domain in DomainNet and obtain the raw weights (before softmax) generated by CMSS. Then, we calculate the number of samples in each domain passing a manually selected threshold $\tau$. We use the number of samples passing this threshold in each domain to indicate the domain preference level. The larger the fraction, more weights are given to samples from the domains, hence, higher the domain preference.
Figure \ref{fig:w_selection_all} shows the visualization of domain preference for each target domain. We picked $3$ different $\tau$ in each experiment for more precise observation. We observe that CMSS does display domain preference (\textit{Clipart} - \textit{Painting}, \textit{Infograph} - \textit{Sketch}, \textit{Real} - \textit{Clipart}) that is in fact correlated with the visual similarity of the domains. An exception is \textit{Quickdraw}, where no domain preference is observed. We argue that this is because \textit{Quickdraw} has significant domain shift compared to all other domains, hence no specific domain is preferred. However, CMSS still produces better performance on \textit{Quickdraw}. While there is no domain preference for \textit{Quickdraw}, there is within-domain sample preference as illustrated in Figure~\ref{fig:w_ranked_clock}. That is, our approach chooses samples within a domain that are structurally more similar to the target domain of interest. Hence, just visualizing aggregate domain preference does not depict the complete picture. We will present sample-wise visualization in the next section.


\begin{figure*}[t]
\centering
    \includegraphics[width=1.0\textwidth]{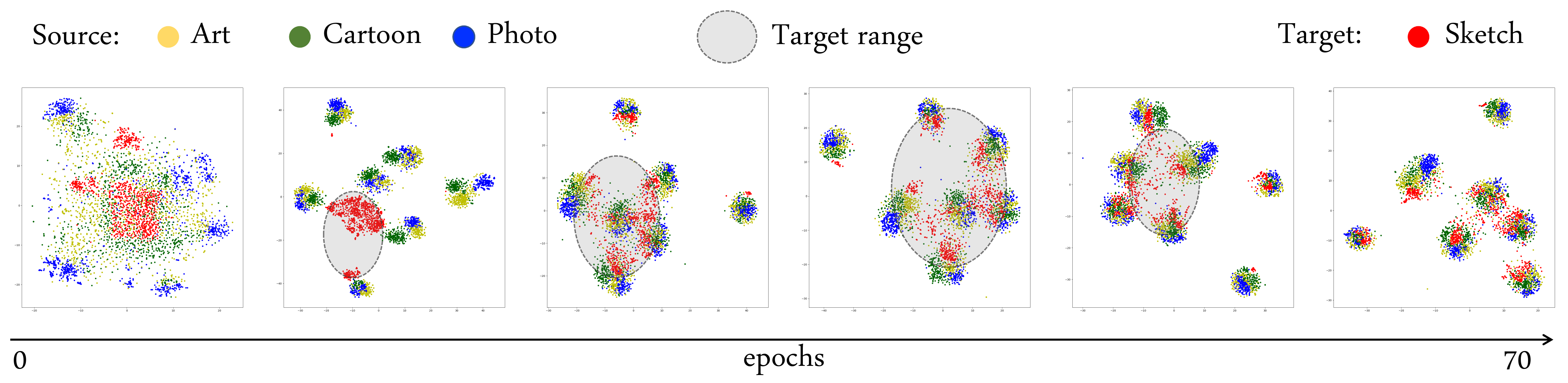}
    \caption{t-SNE visualization of features at six different epochs during training. The shaded region is the migrated range of target features. Dateset used is PACS with \textit{sketch} as the target domain.}
    \label{fig:tsne_over_time}
\end{figure*}

\subsubsection{Beyond Domain Preference} 
In addition to domain preference, we are interested in taking a closer look at sample-wise source selection. To do this, we first obtain the weights generated by CMSS for all source samples and rank the source images according to their weights. An example is shown in Figure \ref{fig:w_ranked_clock}. For better understanding, we visualize samples belonging to a fixed category (``Clock'' in Figure \ref{fig:w_ranked_clock}). See \textit{Appendix} for more visualizations.

In Figure \ref{fig:w_ranked_clock}, we find that notion of similarity discovered by CMSS is different for different domains. When the target domain is \textit{Clipart} (left panel of Figure \ref{fig:w_ranked_clock}), source samples with colors and cartoonish shapes are ranked at the top, while samples with white background and simplistic shapes are ranked at the bottom. When the target is \textit{Quickdraw} (right panel of Figure \ref{fig:w_ranked_clock}), one would think that CMSS will simply be selecting images with similar white background. Instead, it prefers samples which are structurally similar to the regular rounded clock shape (as most samples in \textit{Quickdraw} are similar to these). It thus appears that structural similarity is favored in \textit{Quickdraw}, whereas color information is preferred in \textit{Clipart}. This provides support that CMSS selects samples according to ease of alignment to the target distribution, which is automatically discovered per domain. We argue that this property of CMSS has an advantage over approaches such as MDAN \cite{zhao2018adversarial} which simply weighs manually partitioned domains. 

\subsection{Selection Over Time}
\label{sec:selection_over_time}

In this section, we discuss how source selection varies as training progresses. In Figure \ref{fig:w_variance}, we plot mean and variance of weights (output of Curriculum Manager) over training iterations. We observe that the variance is high initially, which indicates many samples have weights away from the mean value of $1$. Samples with higher weights are preferred, while those with low weights contribute less to the alignment. In the later stages, the variance is very low which indicates most of the weights are close to $1$. Hence, our approach gradually adapts to increasingly many source samples over time, naturally learning a curriculum for adaptation. In Figure \ref{fig:tsne_over_time}, we plot a t-SNE visualization of features at different epochs. We observe that the target domain \textit{sketch} (red) first adapts to \textit{Art} (yellow), and then gradually aligns with \textit{Cartoon} (green) and \textit{Photo} (blue).

\section{Conclusion}
In this paper, we proposed Curriculum Manager for Source Selection (CMSS) that learns a curriculum for Multi-Source Unsupervised Domain Adaptation. A curriculum is learnt that iteratively favors source samples that align better with the target distribution over the entire training. The curriculum learning is achieved by an adversarial interplay with the discriminator, and achieves state-of-the-art on four benchmark datasets. We also shed light on the inner workings of CMSS, and we hope that will pave the way for further advances to be made in this research area.

\section*{Acknowledgement}
This work was supported by Facebook AI Research and DARPA via ARO contract number W911NF2020009.

\bibliographystyle{splncs04}
\bibliography{egbib}

\begin{thebibliography}{10}
\providecommand{\url}[1]{\texttt{#1}}
\providecommand{\urlprefix}{URL }
\providecommand{\doi}[1]{https://doi.org/#1}

\bibitem{ben2010theory}
Ben-David, S., Blitzer, J., Crammer, K., Kulesza, A., Pereira, F., Vaughan,
  J.W.: A theory of learning from different domains. Machine learning
  \textbf{79}(1-2),  151--175 (2010)

\bibitem{ben2007analysis}
Ben-David, S., Blitzer, J., Crammer, K., Pereira, F.: Analysis of
  representations for domain adaptation. In: Advances in neural information
  processing systems. pp. 137--144 (2007)

\bibitem{NIPS2006_2983}
Ben-David, S., Blitzer, J., Crammer, K., Pereira, F.: Analysis of
  representations for domain adaptation. In: Sch\"{o}lkopf, B., Platt, J.C.,
  Hoffman, T. (eds.) Advances in Neural Information Processing Systems 19, pp.
  137--144. MIT Press (2007)

\bibitem{weston2009curriculum}
Bengio, Y., Louradour, J., Collobert, R., Weston, J.: Curriculum learning. In:
  Proceedings of the 26th Annual International Conference on Machine Learning.
  p. 41–48. ICML ’09, Association for Computing Machinery, New York, NY,
  USA (2009)

\bibitem{bhatt2016multi}
Bhatt, H.S., Rajkumar, A., Roy, S.: Multi-source iterative adaptation for
  cross-domain classification. In: IJCAI. pp. 3691--3697 (2016)

\bibitem{NIPS2007_3212}
Blitzer, J., Crammer, K., Kulesza, A., Pereira, F., Wortman, J.: Learning
  bounds for domain adaptation. In: Platt, J.C., Koller, D., Singer, Y.,
  Roweis, S.T. (eds.) Advances in Neural Information Processing Systems 20, pp.
  129--136. Curran Associates, Inc. (2008)

\bibitem{cao2018partial}
Cao, Z., Long, M., Wang, J., Jordan, M.I.: Partial transfer learning with
  selective adversarial networks. In: Proceedings of the IEEE Conference on
  Computer Vision and Pattern Recognition. pp. 2724--2732 (2018)

\bibitem{chen2019progressive}
Chen, C., Xie, W., Huang, W., Rong, Y., Ding, X., Huang, Y., Xu, T., Huang, J.:
  Progressive feature alignment for unsupervised domain adaptation. In:
  Proceedings of the IEEE Conference on Computer Vision and Pattern
  Recognition. pp. 627--636 (2019)

\bibitem{chen2018re}
Chen, Q., Liu, Y., Wang, Z., Wassell, I., Chetty, K.: Re-weighted adversarial
  adaptation network for unsupervised domain adaptation. In: Proceedings of the
  IEEE Conference on Computer Vision and Pattern Recognition. pp. 7976--7985
  (2018)

\bibitem{chen2018domain}
Chen, Y., Li, W., Sakaridis, C., Dai, D., Van~Gool, L.: Domain adaptive faster
  r-cnn for object detection in the wild. In: Proceedings of the IEEE
  conference on computer vision and pattern recognition. pp. 3339--3348 (2018)

\bibitem{ding2018semi}
Ding, Z., Nasrabadi, N.M., Fu, Y.: Semi-supervised deep domain adaptation via
  coupled neural networks. IEEE Transactions on Image Processing
  \textbf{27}(11),  5214--5224 (2018)

\bibitem{duan2009domain}
Duan, L., Tsang, I.W., Xu, D., Chua, T.S.: Domain adaptation from multiple
  sources via auxiliary classifiers. In: Proceedings of the 26th Annual
  International Conference on Machine Learning. pp. 289--296. ACM (2009)

\bibitem{duan2012exploiting}
Duan, L., Xu, D., Chang, S.F.: Exploiting web images for event recognition in
  consumer videos: A multiple source domain adaptation approach. In: 2012 IEEE
  Conference on Computer Vision and Pattern Recognition. pp. 1338--1345. IEEE
  (2012)

\bibitem{duan2012domain}
Duan, L., Xu, D., Tsang, I.W.H.: Domain adaptation from multiple sources: A
  domain-dependent regularization approach. IEEE Transactions on Neural
  Networks and Learning Systems  \textbf{23}(3),  504--518 (2012)

\bibitem{ganin2014unsupervised}
Ganin, Y., Lempitsky, V.: Unsupervised domain adaptation by backpropagation.
  arXiv preprint arXiv:1409.7495  (2014)

\bibitem{gong2012geodesic}
Gong, B., Shi, Y., Sha, F., Grauman, K.: Geodesic flow kernel for unsupervised
  domain adaptation. In: 2012 IEEE Conference on Computer Vision and Pattern
  Recognition. pp. 2066--2073. IEEE (2012)

\bibitem{jeong2019self}
Jeong, R., Aytar, Y., Khosid, D., Zhou, Y., Kay, J., Lampe, T., Bousmalis, K.,
  Nori, F.: Self-supervised sim-to-real adaptation for visual robotic
  manipulation. arXiv preprint arXiv:1910.09470  (2019)

\bibitem{kang2019contrastive}
Kang, G., Jiang, L., Yang, Y., Hauptmann, A.G.: Contrastive adaptation network
  for unsupervised domain adaptation. In: Proceedings of the IEEE Conference on
  Computer Vision and Pattern Recognition. pp. 4893--4902 (2019)

\bibitem{kumar2018co}
Kumar, A., Sattigeri, P., Wadhawan, K., Karlinsky, L., Feris, R., Freeman, B.,
  Wornell, G.: Co-regularized alignment for unsupervised domain adaptation. In:
  Advances in Neural Information Processing Systems. pp. 9345--9356 (2018)

\bibitem{kurmi2019curriculum}
Kurmi, V.K., Bajaj, V., Subramanian, V.K., Namboodiri, V.P.: Curriculum based
  dropout discriminator for domain adaptation. arXiv preprint arXiv:1907.10628
  (2019)

\bibitem{lee2019drop}
Lee, S., Kim, D., Kim, N., Jeong, S.G.: Drop to adapt: Learning discriminative
  features for unsupervised domain adaptation. In: Proceedings of the IEEE
  International Conference on Computer Vision. pp. 91--100 (2019)

\bibitem{li2017deeper}
Li, D., Yang, Y., Song, Y.Z., Hospedales, T.M.: Deeper, broader and artier
  domain generalization. In: Proceedings of the IEEE International Conference
  on Computer Vision. pp. 5542--5550 (2017)

\bibitem{li2018deep}
Li, Y., Tian, X., Gong, M., Liu, Y., Liu, T., Zhang, K., Tao, D.: Deep domain
  generalization via conditional invariant adversarial networks. In:
  Proceedings of the European Conference on Computer Vision. pp. 624--639
  (2018)

\bibitem{liu2019transferable}
Liu, H., Long, M., Wang, J., Jordan, M.: Transferable adversarial training: A
  general approach to adapting deep classifiers. In: International Conference
  on Machine Learning. pp. 4013--4022 (2019)

\bibitem{luo2019taking}
Luo, Y., Zheng, L., Guan, T., Yu, J., Yang, Y.: Taking a closer look at domain
  shift: Category-level adversaries for semantics consistent domain adaptation.
  In: Proceedings of the IEEE Conference on Computer Vision and Pattern
  Recognition. pp. 2507--2516 (2019)

\bibitem{mancini2019adagraph}
Mancini, M., Bul{\`o}, S.R., Caputo, B., Ricci, E.: Adagraph: Unifying
  predictive and continuous domain adaptation through graphs. In: Proceedings
  of the IEEE Conference on Computer Vision and Pattern Recognition. pp.
  6568--6577 (2019)

\bibitem{mancini2018boosting}
Mancini, M., Porzi, L., Rota~Bul{\`o}, S., Caputo, B., Ricci, E.: Boosting
  domain adaptation by discovering latent domains. In: Proceedings of the IEEE
  Conference on Computer Vision and Pattern Recognition. pp. 3771--3780 (2018)

\bibitem{motiian2017few}
Motiian, S., Jones, Q., Iranmanesh, S., Doretto, G.: Few-shot adversarial
  domain adaptation. In: Advances in Neural Information Processing Systems. pp.
  6670--6680 (2017)

\bibitem{ouyang2019data}
Ouyang, C., Kamnitsas, K., Biffi, C., Duan, J., Rueckert, D.: Data efficient
  unsupervised domain adaptation for cross-modality image segmentation. In:
  International Conference on Medical Image Computing and Computer-Assisted
  Intervention. pp. 669--677. Springer (2019)

\bibitem{pan2019transferrable}
Pan, Y., Yao, T., Li, Y., Wang, Y., Ngo, C.W., Mei, T.: Transferrable
  prototypical networks for unsupervised domain adaptation. In: Proceedings of
  the IEEE Conference on Computer Vision and Pattern Recognition. pp.
  2239--2247 (2019)

\bibitem{pei2018multi}
Pei, Z., Cao, Z., Long, M., Wang, J.: Multi-adversarial domain adaptation. In:
  Thirty-Second AAAI Conference on Artificial Intelligence (2018)

\bibitem{peng2018moment}
Peng, X., Bai, Q., Xia, X., Huang, Z., Saenko, K., Wang, B.: Moment matching
  for multi-source domain adaptation. arXiv preprint arXiv:1812.01754  (2018)

\bibitem{saito2018maximum}
Saito, K., Watanabe, K., Ushiku, Y., Harada, T.: Maximum classifier discrepancy
  for unsupervised domain adaptation. In: Proceedings of the IEEE Conference on
  Computer Vision and Pattern Recognition. pp. 3723--3732 (2018)

\bibitem{shu2019transferable}
Shu, Y., Cao, Z., Long, M., Wang, J.: Transferable curriculum for
  weakly-supervised domain adaptation. In: Proceedings of the AAAI Conference
  on Artificial Intelligence. vol.~33, pp. 4951--4958 (2019)

\bibitem{sun2019unsupervised}
Sun, Y., Tzeng, E., Darrell, T., Efros, A.A.: Unsupervised domain adaptation
  through self-supervision. arXiv preprint arXiv:1909.11825  (2019)

\bibitem{valada2019self}
Valada, A., Mohan, R., Burgard, W.: Self-supervised model adaptation for
  multimodal semantic segmentation. International Journal of Computer Vision
  pp. 1--47 (2019)

\bibitem{wang2019few}
Wang, T., Zhang, X., Yuan, L., Feng, J.: Few-shot adaptive faster r-cnn. In:
  Proceedings of the IEEE Conference on Computer Vision and Pattern
  Recognition. pp. 7173--7182 (2019)

\bibitem{wu2019ace}
Wu, Z., Wang, X., Gonzalez, J.E., Goldstein, T., Davis, L.S.: Ace: Adapting to
  changing environments for semantic segmentation. arXiv preprint
  arXiv:1904.06268  (2019)

\bibitem{xiong2014latent}
Xiong, C., McCloskey, S., Hsieh, S.H., Corso, J.J.: Latent domains modeling for
  visual domain adaptation. In: Twenty-Eighth AAAI Conference on Artificial
  Intelligence (2014)

\bibitem{xu2018deep}
Xu, R., Chen, Z., Zuo, W., Yan, J., Lin, L.: Deep cocktail network:
  Multi-source unsupervised domain adaptation with category shift. In:
  Proceedings of the IEEE Conference on Computer Vision and Pattern
  Recognition. pp. 3964--3973 (2018)

\bibitem{yoon2019self}
Yoon, J.S., Shiratori, T., Yu, S.I., Park, H.S.: Self-supervised adaptation of
  high-fidelity face models for monocular performance tracking. In: Proceedings
  of the IEEE Conference on Computer Vision and Pattern Recognition. pp.
  4601--4609 (2019)

\bibitem{you2019towards}
You, K., Wang, X., Long, M., Jordan, M.: Towards accurate model selection in
  deep unsupervised domain adaptation. In: International Conference on Machine
  Learning. pp. 7124--7133 (2019)

\bibitem{zhang2018importance}
Zhang, J., Ding, Z., Li, W., Ogunbona, P.: Importance weighted adversarial nets
  for partial domain adaptation. In: Proceedings of the IEEE Conference on
  Computer Vision and Pattern Recognition. pp. 8156--8164 (2018)

\bibitem{zhang2019few}
Zhang, J., Chen, Z., Huang, J., Lin, L., Zhang, D.: Few-shot structured domain
  adaptation for virtual-to-real scene parsing. In: Proceedings of the IEEE
  International Conference on Computer Vision Workshops. pp.~0--0 (2019)

\bibitem{zhao2019learning}
Zhao, H., Combes, R.T.d., Zhang, K., Gordon, G.J.: On learning invariant
  representation for domain adaptation. arXiv preprint arXiv:1901.09453  (2019)

\bibitem{zhao2018adversarial}
Zhao, H., Zhang, S., Wu, G., Moura, J.M., Costeira, J.P., Gordon, G.J.:
  Adversarial multiple source domain adaptation. In: Advances in Neural
  Information Processing Systems. pp. 8559--8570 (2018)

\end{thebibliography}
\end{document}